\pdfoutput=1
\documentclass[11pt]{article}

\usepackage[]{EMNLP2022}

\usepackage{times}
\usepackage{latexsym}

\usepackage[T1]{fontenc}
\usepackage[utf8]{inputenc}

\usepackage{microtype}

\usepackage{hyperref}       % hyperlinks
\usepackage{url}            % simple URL typesetting
\usepackage{booktabs}       % professional-quality tables
\usepackage{amsfonts}       % blackboard math symbols
\usepackage{multirow}
\usepackage{lipsum}
\usepackage{indentfirst}
\usepackage{amsmath}
\usepackage{dsfont}
\usepackage{relsize}
\usepackage{svg}
\usepackage{amsthm}
\usepackage{xspace}
\usepackage{caption}
\usepackage{subcaption}
\usepackage{listings}
\usepackage{threeparttable}
\usepackage{array}
\usepackage{float}
\usepackage{ragged2e}
\usepackage{xcolor}
\usepackage{amssymb}
\usepackage{dblfloatfix}

\theoremstyle{definition}
\newtheorem{definition}{Definition}

\newcommand{\red}[1]{\textcolor{red}{#1}}

\newcommand{\tool}{{GraphQ IR}\xspace}
\newcommand{\toolx}{{GraphQ IR*}\xspace}
\newcommand{\tools}{{GraphQ IR's}\xspace}
\newcommand{\eg}{\textit{e.g.},\xspace}
\newcommand{\ie}{\textit{i.e.},\xspace}

\title{GraphQ IR: Unifying the Semantic Parsing of Graph Query Languages \\with One Intermediate Representation}

\author{
    \textbf{Lunyiu Nie}$^{1}$, \textbf{Shulin Cao}$^{1}$, \textbf{Jiaxin Shi}$^{2}$, \textbf{Jiuding Sun}$^{1}$, \\
    \textbf{Qi Tian}$^{2}$, \textbf{Lei Hou}$^{1}$, \textbf{Juanzi Li}$^{1}$,  \textbf{Jidong Zhai}$^{1}$ \\ \\
    $^1$ Department of Computer Science and Technology, Tsinghua University \\ 
    $^2$ Huawei Cloud Computing Technologies Co., Ltd. \\
    \texttt{\{nlx20, caosl19, sjd22\}@mails.tsinghua.edu.cn},\hspace{.5em}\texttt{shijx12@gmail.com}, \\
    \texttt{tian.qi1@huawei.com},\hspace{.5em}\texttt{\{houlei,lijuanzi,zhaijidong\}@tsinghua.edu.cn}
}    

\begin{document}
\maketitle
\begin{abstract}
Subject to the huge semantic gap between natural and formal languages, neural semantic parsing is typically bottlenecked by its complexity of dealing with both input semantics and output syntax. Recent works have proposed several forms of supplementary supervision but none is generalized across multiple formal languages. This paper proposes a unified intermediate representation (IR) for graph query languages, named \tool. It has a natural-language-like expression that bridges the semantic gap and formally defined syntax that maintains the graph structure. Therefore, a neural semantic parser can more precisely convert user queries into \tool, which can be later losslessly compiled into various downstream graph query languages. Extensive experiments on several benchmarks including \textsc{Kqa Pro}, \textsc{Overnight}, \textsc{GrailQA} and \textsc{MetaQA}-Cypher under standard i.i.d., out-of-distribution and low-resource settings validate \tool's superiority over the previous state-of-the-arts with a maximum 11\% accuracy improvement. 
\end{abstract}

\section{Introduction}
By mapping natural language utterances to logical forms, the task of semantic parsing has been widely explored in various applications, including database query \cite{yu2018spider, talmor2018web} and general-purpose code generation \cite{yin2017syntactic, campagna2019genie, nan2020hisyn}. Although the methodology has evolved from earlier statistical approaches \cite{zettlemoyer2005learning, kwiatkowksi2010inducing} to present Seq2Seq paradigm \cite{zhong2017seq2sql, damonte2021one}, the semantic gap between natural language and logical forms still lies as the major challenge for semantic parsing. 

\begin{figure}[t]
    \centering
    \resizebox{\linewidth}{!}{
    \includegraphics{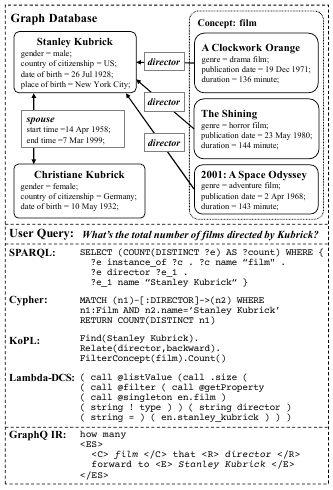}
    }
    \caption{A a property graph extracted from Wikidata \cite{vrandevcic2014wikidata}. We present a relevant user query with its corresponding logical forms in different query languages and in \tool. }
    \label{fig:examples}
\end{figure}

As shown in Figure \ref{fig:examples}, in graph query languages (\eg SPARQL, Cypher, Lambda-DCS, and newly emerged KoPL, etc.), graph nodes, edges and their respective properties constitute the key semantics of the logical forms \cite{perez2009semantics}, which are very different from the expression of natural language utterances. Such discrepancy significantly hinders the learning of neural semantic parsers and therefore increases the demand for labeled data \cite{yin2021ingredients}. However, due to the laborious efforts and language-specific expertise required in annotation, such demand cannot always be satisfied and thus becomes the bottleneck \cite{li2020context, herzig2021unlocking}.

To overcome these challenges, many works adopt complementary forms of supervision, such as the schema of database \cite{hwang2019comprehensive}, results of the execution \cite{clarke2010driving, wang2018robust, wang2021learning}, and grammar-constrained decoding algorithms \cite{krishnamurthy2017neural, shin2021constrained, baranowski2021grammar}. Although effective, the additional resources that these methods rely on are not necessarily available in practice. By normalizing the expression \cite{berant2014semantic, su2017cross} or enriching the structure \cite{reddy2016transforming, cheng2017learning, hu2018state} of natural language utterances, another category of works proposes various intermediate representations like AMR \cite{kapanipathi2021leveraging} to ease the parsing of complex queries. However, the transition from their IRs to the downstream logical forms may incur extra losses in precision \cite{bornea2021learning}. Besides, these representations are usually coupled to specific data or logical forms and thus cannot be easily transferred to other tasks or languages \cite{kamath2018survey}. 

In industry, aside from SPARQL, many other graph query languages such as Cypher \cite{francis2018cypher} and Gremlin \cite{rodriguez2015gremlin} are equally or even more commonly used in graph database interaction \cite{angles2012comparison, seifer2019empirical}. However, most graph query semantic parsing works only support SPARQL \cite{talmor2018web, dubey2019lc, keysers2019measuring} while very few works target other graph query languages. Meanwhile, no existing tools or IR can support the data conversion among multiple graph query languages \cite{moreira2020sparqling, a2022sparql}. Such lack of interoperability has not only hindered the semantic parsing of low-resource languages but also limited the potential of querying heterogeneous databases \cite{mami2019query, DBLP:conf/amw/AnglesTT19}.

In this paper, we propose a unified intermediate representation for graph query languages, namely \tool, to resolve these issues from a novel perspective. The designs of \tool weigh up the semantics of both natural and formal language by (a) producing the IR sequences with composition rules consistent with modern English \cite{tomlin2014basic} to close the semantic gap; and (b) maintaining the fundamental graph structures like nodes, edges, and properties, such that the IR can be automatically compiled into any downstream graph query languages without any loss. 

Instead of directly mapping the user query to the logical form, we first parse natural language into \tool, then compile the IR into the target graph query languages (\eg SPARQL, Cypher, Lambda-DCS, KoPL, etc.). Therefore, language-specific grammar features that initially posed a huge obstacle to semantic parsing are now explicitly handled by the compiler. Additionally, with the \tool as a bridge, our implemented source-to-source compiler can support lossless translation among multiple graph query languages and thus unify the annotations of different languages for eliminating the data bottleneck.

To validate the effectiveness of \tool, we conducted extensive experiments on benchmarks \textsc{Kqa Pro}, \textsc{Overnight}, \textsc{GrailQA} and \textsc{MetaQA}-Cypher. Results show that our approach can consistently outperform the previous works by a significant margin. Especially under the compositional and few-shot generalization settings, our approach with \tool can demonstrate a maximum 11\% increase in accuracy over the baselines.
% By computing and visualizing the semantic distance between natural language and IR, we also affirm that \tool can indeed close the semantic gap effectively.

The main contributions of our work include:
\vspace{-\topsep}
\vspace{0.3em}
\begin{itemize}
\itemsep0em 
    \item We propose \tool for unifying the semantic parsing of graph query languages and present the IR design principles that are critical for bridging the semantic gap; 
    \item Experimental results show that our approach can consistently achieve state-of-the-art performance across multiple benchmarks under the standard i.i.d, out-of-distribution, and low-resource settings. 
    \item Our implemented source-to-source compiler unlocks data interoperability by supporting the bi-directional translation among different graph query languages. The code and toolkit are publicly available at \url{https://github.com/Flitternie/GraphQ_IR}. 
\end{itemize}
\section{\tool}
In this section, we formalize the grammar and the expressiveness of our \tool based on the definition of property graph and regular path query. Then we summarize the design principles of \tool for bridging the semantic gap between natural and formal language as well as unifying different graph query languages.

\subsection{Definition}
\label{sec:grammar}
As the top of Figure \ref{fig:examples} demonstrates, a graph database can be expressed as a collection of property graphs that include \textbf{Entity} (graph nodes, \eg Stanley Kubrick), \textbf{Attribute} (node properties, \eg date of birth), \textbf{Concept} (node label, \eg film), \textbf{Relationship} (graph edges, \eg spouse) and \textbf{Qualifier} (edge properties, \eg start time). 

Therefore, to evaluate the expressiveness of \tool, we start by giving the definition of property graph: a directed labeled multigraph where each node or edge can contain a set of property-value pairs \cite{DBLP:conf/amw/Angles18}. 

\begin{definition}[\textbf{\textit{Property graph}}]
A property graph $G$ is a tuple ($N$, $E$, $\rho$, $\lambda$, $\sigma$) where:\\
(1) $N$ is a finite set of \textit{nodes}. \\
(2) $E$ is a finite set of \textit{edges} such that $N \cap E= \varnothing$. \\
(3) $\rho: E \rightarrow (N \times N)$ is a total function. Specifically, $\rho(e) = (n_1, n_2)$ refers $e$ is a directed edge from node $n_1$ to $n_2$. \\
(4) $\lambda: (N \cup E) \rightarrow L$ is a partial function where $L$ is a set of labels. Specifically, if $\lambda(n)=l$ then $l$ is the label of node $n$. \\
(5) $\sigma: (N \cup E) \times P \rightarrow V$ is a partial function with $P$ a set of properties and $V$ a set of values $V$. Specifically, if $\sigma(n, p)=v$ then the property $p$ of node $n$ has value $v$. 
\end{definition}

Subsequently, a graph path can be expressed as $\pi=(n_1, e_1, ..., e_{k-1}, n_k)$ where $k \geq 1$ with each $e_i$ being the edge between $n_i$ and $n_{i+1}$. The spelling of path, denoted as $\lambda(\pi)$, is the concatenation of edge labels $\lambda(e_1)...\lambda(e_{k-1})$ \cite{mendelzon1995finding, barcelo2013querying}.

\begin{definition}[\textbf{\textit{Regular path query}}]
A regular path query has the general form $Q=x \xrightarrow{\alpha} y$ where x denotes the start point, $\alpha$ is a regular expression defined over $\lambda(\pi)$, and $y$ denotes the endpoints of the query.
\end{definition}

By incorporating $\rho$, $\lambda$, $\sigma$ and their inverse function $\rho^{-1}$, $\lambda^{-1}$, $\sigma^{-1}$, such regular path query can be extended to support navigational queries towards any graph elements $\psi \in (N\cup E\cup L \cup P \cup V)$ \cite{wood2012query, van2016pgql}. We can now evaluate the expressiveness of a language.

\begin{definition}[\textbf{\textit{Path query expressiveness}}] A path query ${q}$ is expressible in a language $\mathcal{L}$, if there exists an expression $\varepsilon \in \mathcal{L}$ such that, for any subgraph $G' \subseteq G$, we have $\varepsilon(G') = {q}(G')$ \cite{fletcher2015relative}.
\end{definition}

We formalize \tool as a context-free grammar $(\mathcal{V}, \Sigma, \mathcal{S}, \mathcal{P})$ and present its non-terminals and productions in Appendix Table \ref{tab:full_grammar}. Its $\mathcal{V}$ and $\mathcal{P}$ are respectively defined as the superset of the terminal set ($n$, $e$, $l$, $p$, $v$) and production set ($\rho$, $\lambda$, $\sigma$, $\rho^{-1}$,  $\lambda^{-1}$, $\sigma^{-1}$) of regular graph query. Therefore, all path queries expressible in regular grammar are also expressible in the context-free grammar of \tool \cite{hopcroft2006automata}. Furthermore, \tool also supports extended operations like \textit{Union}, \textit{Difference} and \textit{Filter} to express complex graph query patterns \cite{angles2017foundations}. 

Empirically, \tool can express all graph query patterns that appeared in benchmarks \textsc{Kqa Pro}, \textsc{Overnight}, \textsc{GrailQA} and \textsc{MetaQA}-Cypher, with details elaborated in Section \ref{sec:dataset}.

% Therefore, a basic graph query pattern can be expressed as a subgraph $G'$ where its component sets (\ie $N', E', L', P', V'$) may contain unknown variables other than specified constants \cite{barcelo2013querying, angles2017foundations}. Accordingly, we formalize the context-free-grammar of \tool by defining the constant elements as terminal symbols and the variables as the non-terminals with production rules derived from $\rho$, $\lambda$ and $\sigma$. A subset of \tools grammar is presented in Appendix Table \ref{tab:full_grammar}.

\subsection{Principles}
We summarize several principles in designing \tool in this way: present in a syntax close to natural language while preserving the structural semantics equivalent to formal languages.  

\subsubsection{Diminishing syntactical discrepancy}
To facilitate the training of the neural semantic parser, the target IR sequence should share a similar syntax in correspondence to the input utterance. 

To achieve this, the IR structure should first match how users typically raise queries. Therefore, we simplify the triple-based structure in graph query languages into a more natural subject-verb-object syntactic construction \cite{tomlin2014basic}. Take Figure \ref{fig:examples}'s task setting as an example, the two triples \texttt{(?e instance\_of ?c)} and \texttt{(?c name ``film'')} as the entity concept constraint in SPARQL are simplified to the sentence subject ``\texttt{<C>} \textit{film} \texttt{</C>}'' in \tool. Multi-hop relationship and attribute queries are formulated as relative clauses similar to the English expression and thus can be comfortably generated by a language-model-based neural semantic parser.   

Secondly, IR should also leave out the variables (\eg \texttt{?e}, \texttt{?c} in SPARQL) and operators (\eg \texttt{SELECT}, \texttt{WHERE}, \texttt{RETURN}, etc.) in logical forms that cannot be easily aligned to natural language utterances. Alternatively, human-readable operators are adopted in \tool, as illustrated in Appendix Table \ref{tab:full_grammar}.

\begin{figure*}[ht!]
     \centering
     \includegraphics[width=\textwidth]{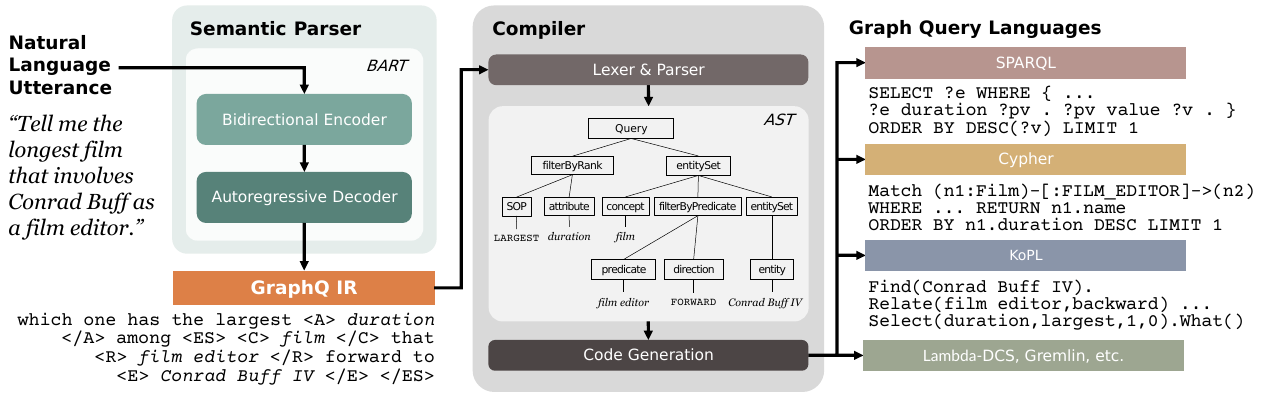}
     \caption{Overall implementation of our proposed framework. The user queries are first converted to \tool sequences by a semantic parser and subsequently transpiled into the target graph query languages by a compiler. } 
     \label{fig:main}
\end{figure*}

\subsubsection{Eliminating semantic ambiguity}
In formal languages, multiple parallel implementations can achieve the same functionalities. However, such redundancy and ambiguity in semantics may pose challenges to the neural semantic parser. 

For example, in Lambda-DCS, there co-exist three implementations for constraining one's concept (\eg Kobe is a \textit{player}), respectively through:
\vspace{-\topsep}
\begin{itemize}
\setlength{\parskip}{0pt} \setlength{\itemsep}{0pt plus 1pt}
    \item \textbf{\textit{EventNP}}:\texttt{(call @getProperty (en.player.kobe\_bryant) (call @reverse (string player)))}; 
    \item \textbf{\textit{TypeNP}}:\texttt{(call @getProperty (call @singleton ( en.player )) (string !type))}; and 
    \item \textbf{\textit{DomainNP}}:\texttt{(call @getProperty ( en.player.kobe\_bryant ... (call @domain (string player))) (string player))}.
\end{itemize}

When designing an IR, such redundant and ambiguous semantics should be clarified into more definitive and orthogonal representations \cite{campagna2019genie}. Thus in \tool, we unify all such unnecessary distinctions and prune redundant structures in logical forms to distill the core semantics. In the previous example, \tool only requires a simple noun modifier ``\texttt{<C>} \textit{player} \texttt{</C>}'' as the concept constraint. This not only makes the language clearer for users and semantic parsers to comprehend, but also facilitates the next-step compilation from the IR to the downstream formal language.

\subsubsection{Maintaining graph structural semantics}
In addition to the aforementioned designs to improve alignment with natural language, the syntax of IR also needs to maintain the key structures of graph queries for subsequent lossless compilation. 

Specifically, IR should keep track of the data types of graph structural elements. We design \tool to be strong-typing by explicitly stating the type of terminal nodes with respective special tokens, \eg \texttt{<E>} for \texttt{Entity}, \texttt{<R>} for \texttt{Relation}, \texttt{<A>} for \texttt{Attribute}, etc. Values of different types are also differentiated in \tool with our pre-defined or user custom indicators, \eg \texttt{string}, \texttt{number}, \texttt{date}, \texttt{time}, etc. 

Furthermore, IR should also preserve the hierarchical dependencies that are critical for multi-hop queries. We introduce \texttt{<ES>} as a scoping token in \tool to explicitly indicate the underlying dependencies among the clauses produced by an \texttt{EntitySet}, as shown in Appendix Table \ref{tab:full_grammar}. Such scoping tokens in \tool can facilitate the compiler to recover the hierarchical structure and finally convert the IR sequences into one of the graph query languages deterministically.
\section{Implementation}
\label{sec:imple}
We depict the full picture of our proposed framework in Figure \ref{fig:main}. The neural semantic parser first maps the input natural language utterance into \tool. Thereafter, the \tool sequence is fed into the compiler and parsed into an abstract syntax tree for downstream graph query language code generation.

\subsection{Neural Semantic Parser}
\label{sec:parser}
To verify the above principles in practice, we formulate the conversion from natural language to our \tool as a Seq2Seq task and adopt an encoder-decoder framework for implementing the neural semantic parser. 

As shown in the left part of Figure \ref{fig:main}, the encoder module of the semantic parser first maps the input natural language utterance $X$ to a high dimensional feature space with non-linear transformations for capturing the semantics of the input tokens. The decoder module subsequently then interprets the hidden representations and generates the IR sequence by factorizing the probability distribution:
\begin{equation}
    p(\textit{IR}) = \prod_{i=1}^{n} P(y_i|X,y_1,...,y_{i-1}),
\end{equation}
where $y_i$ is the $i$-th token of IR sequence with in total $n$ tokens. Specifically, we implement this encoder-decoder network with BART \cite{lewis2020bart}, a pretrained language model that is proficient in comprehending the diverse user utterances and generating the \tool sequences that are structured in natural-language-like expressions.

Please note that the implementation in this part is orthogonal to our \tool and can be substituted by other semantic parsing models. 

\subsection{Compiler}
\label{sec:compiler}
The implementation of \tools compiler comprises a front-end module that generates an abstract syntax tree from the IR sequence and a back-end module that transforms the tree structure into the target graph query language. 

% Following the grammar we defined in Section \ref{sec:grammar}, the compiler front-end first performs lexical and syntax analysis on the IR sequence. By formalizing \tools lexer and parser rules in extended Backus–Naur form, we use ANTLR \cite{parr2013definitive} to generate an LL(*) parser that performs the leftmost derivation of a sentence with tokens look-ahead \cite{parr2011ll}. Such parser can automatically parse the sequence of \tool into an abstract syntax tree that contains syntactic dependencies and hierarchical structures. 

The compiler front-end is responsible for performing the lexical and syntax analysis on the IR sequence. The lexer first splits the sequence into lexical tokens, which are subsequently structured into a parse tree with \textit{LL(*)} parsing strategy \cite{parr2013definitive} according to the pre-defined grammar in Section \ref{sec:grammar}. As such, \tool sequence can be automatically constructed into an abstract syntax tree (AST) that contains syntactic dependencies and hierarchical structures. 

The compiler back-end will then traverse the abstract syntax tree and restructure the nodes and dependencies into one of the downstream graph query languages. We formalize the code generation as a tree mapping process, where the subtrees carrying equivalent information are aligned according to pre-defined transformation rules. To illustrate, we present 2 examples of generating SPARQL and Lambda-DCS queries respectively in Appendix Figure \ref{fig:ast_2} and Figure \ref{fig:ast_1}.

Similarly, we also implement the compiler that supports conversion from graph query languages to \tool. Thus, with the IR as a middleware, our toolkit can also achieve the transpilation between any two graph query languages supported.  
\section{Experiments}
In this section, we evaluate \tool on several benchmarks under different task settings.

\begin{table*}[ht!]
\centering
\small
\resizebox{\textwidth}{!}{
\begin{tabular}{lcccccccc}
\toprule
 & \textbf{Multi-hop} & \textbf{Qualifier} & \textbf{Comparison} & \textbf{Logical} & \textbf{Count} & \textbf{Verify} & \textbf{Zero-shot} & \textbf{Overall} \\ 
\midrule
\textbf{Baselines} & & \\
RGCN \cite{schlichtkrull2018rgcn} & 34.00 & 27.61 & 30.03 & 35.85 & 41.91 & 65.88 & - & 35.07\\
BART+SPARQL \cite{shi2020kqa} & 88.49 & 83.09 & 96.12 & 88.67 & 85.78 & 92.33 & 87.88 & 89.68 \\
BART+KoPL \cite{shi2020kqa} & 89.46 & 84.76 & 95.51 & 89.30 & 86.68 & 93.30 & 89.59 & 90.55\\
CFQ IR \cite{herzig2021unlocking} & 87.51 & 81.32 & 95.70 & 90.33 & 86.23 & 92.20 & 87.12 & 88.96 \\
\hline
\textbf{Our Approach} & & \\
\tool  & \textbf{90.38} & \textbf{84.90} & \textbf{97.15} & \textbf{92.64} & \textbf{89.39} & \textbf{94.20} & \textbf{94.20} & \textbf{91.70}\\
% \toolx & & \\
\bottomrule
\end{tabular}
}
\caption{Test accuracies on \textsc{Kqa Pro} dataset. Data are categorized into \textsc{Multi-hop} queries with multi-hop inference, \textsc{Qualifier} knowledge queries, \textsc{Comparison} between several entities, \textsc{Logical} union or intersection, \textsc{Count} queries for the quantity of entities, \textsc{Verify} queries with a boolean answer, and \textsc{Zero-shot} queries whose answer is not seen in the training set.}
\label{tab:kqapro}
\end{table*}

\begin{table*}[ht!]
\centering
\resizebox{\textwidth}{!}{
\begin{tabular}{p{0.375\textwidth}>{\centering}p{0.0725\textwidth}>{\centering}p{0.0725\textwidth}>{\centering}p{0.0725\textwidth}>{\centering}p{0.0725\textwidth}>{\centering}p{0.0725\textwidth}>{\centering}p{0.0725\textwidth}>{\centering}p{0.0725\textwidth}>{\centering}p{0.0725\textwidth}>{\centering\arraybackslash}p{0.08\textwidth}}
\toprule
 & \textbf{Bas.} & \textbf{Blo.} & \textbf{Cal.} & \textbf{Hou.} & \textbf{Pub.} & \textbf{Rec.} & \textbf{Res.} & \textbf{Soc.} & \textbf{Overall} \\
\midrule
\textbf{Baselines} \\
SPO \cite{wang2015building} & 46.3 & 41.9 & 74.4 & 54.0 & 59.0 & 70.8 & 75.9 & 48.2 & 58.8 \\
CrossDomain* \cite{su2017cross} & 88.2 & 62.2 & \textbf{82.1} & 78.8 & 80.1 & 86.1 & 83.7 & 83.1 & 80.6 \\ 
Seq2Action \cite{chen2018sequence} & 88.2 & 61.4 & 81.5 & 74.1 & 80.7 & 82.9 & 80.7 & 82.1 & 79.0 \\
DUAL \cite{cao2019semantic} & 84.9 & 61.2 & 78.6 & 67.2 & 78.3 & 80.6 & 78.9 & 81.3 & 76.4 \\
2-stage DUAL* \cite{cao2020unsupervised} & 87.2 & \textbf{65.7} & 80.4 & 75.7 & 80.1 & 86.1 & 82.8 & 82.7 & 80.1 \\
\hline
\textbf{Our Approach} \\
\tool & 88.2 & 64.7 & 78.6 & 72.0 & 77.6 & 83.3 & 84.9 & 81.6 & 79.5 \\
\toolx & \textbf{88.2} & 65.4 & 81.6 & \textbf{81.5} & \textbf{82.6} & \textbf{92.9} & \textbf{89.8} & \textbf{84.1} & \textbf{82.1} \\
\bottomrule
\end{tabular}
}
\caption{Test accuracies on \textsc{Overnight} dataset. Methods with asterisk (*) involve cross-domain training.}
\label{tab:overnight}
\end{table*}

\subsection{Datasets}
\label{sec:dataset}
For evaluation, we test on benchmarks \textsc{Kqa Pro}, \textsc{Overnight}, \textsc{GrailQA} and \textsc{MetaQA}-Cypher that altogether cover graph query languages SPARQL, KoPL, Lambda-DCS, and Cypher. 

In all experiments, the \tool sequences are automatically converted from the original logical forms of the respective datasets by the bi-directional compiler without extra re-annotation.

\paragraph{KQA Pro} \textsc{Kqa Pro} \cite{shi2020kqa} is a large-scale dataset for complex question answering over Wikidata knowledge base \cite{vrandevcic2014wikidata}. It is the largest KBQA corpus that contains 117,790 natural language questions along with the corresponding SPARQL and KoPL logical forms, covering complex graph queries involving multi-hop inference, logical union and intersection, etc. In our experiment, it is divided into 94,376 train, 11,797 validation, and 11,797 test cases.

\paragraph{Overnight} \textsc{Overnight} \cite{wang2015building} is a semantic parsing dataset with 13,682 examples across 8 sub-domains extracted from Freebase \cite{bollacker2008freebase}. Each domain has natural language questions and pairwise Lambda-DCS queries executable on SEMPRE \cite{berant2013semantic}. It exhibits diverse linguistic phenomena and semantic structures across domains, e.g., temporal knowledge in \textsc{Calendar} domain and spatial knowledge in \textsc{Blocks} domain. We use the same train/val/test splits as in the previous work \cite{wang2015building}.   

\paragraph{GrailQA} \textsc{GrailQA} \cite{gu2021beyond} is a knowledge base question answering dataset with 64k questions grounded on Freebase \cite{bollacker2008freebase} that evaluate generalizability at three levels, \ie i.i.d, compositional generalization and zero-shot. To focus on the sole task of semantic parsing, we replace the entity IDs (\eg \texttt{m.06mn7}) with their respective names (\eg \texttt{Stanley Kubrick}) in \textsc{GrailQA}'s logical forms, thus eliminating the need for an explicit entity linking module as in previous works \cite{chen2021retrack, ye2022rng}. Since \textsc{GrailQA}'s test set is not publicly available for such transformation, we report the validation set results for our evaluation, which have been studied to show consistent trends with the test set \cite{DBLP:journals/corr/abs-2204-08109}.

\paragraph{MetaQA-Cypher} \textsc{MetaQA} \cite{zhang2018variational} contains more than 400k multi-hop QA pairs over WikiMovies knowledge base \cite{miller2016key}. Many studies have previously worked on its SPARQL annotation \cite{huang2021unseen}. Instead, we reconstruct \textsc{MetaQA} into Cypher as a few-shot learning benchmark to evaluate the interoperability achieved by \tool. To the best of our knowledge, this is also the first Cypher dataset in the community of semantic parsing. 

\subsection{Metric}
We adopt \textit{execution accuracy} as our metric based on whether the generated logical form queries can return correct answers. For queries with multiple legal answers, we require the execution results to exactly match \textit{all} ground-truth answers.

\subsection{Results}
\paragraph{I.I.D. Generalization} As Table \ref{tab:kqapro} illustrates, on \textsc{Kqa Pro}, our proposed approach with \tool consistently outperforms the previous approaches on all query categories. In particular, \tool exhibits good generalization under the complex \textsc{Multi-hop}, \textsc{Qualifier} and \textsc{Zero-shot} settings with even larger margins over the baselines. We attribute this to its natural-language-like representations that effectively close the semantic gap and its formally-defined syntax that can be losslessly converted into downstream languages.   

As for \textsc{Overnight}, our methods also significantly surpass the baselines as shown in Table \ref{tab:overnight}. Previous works usually train separate parsers for each of the eight domains 
 due to their distinct vocabularies and grammars \cite{wang2015building, chen2018sequence}. With an extra layer of \tool for unification, domain-specific data are now consolidated into one universal representation, and the training of one domain can thereby benefit from the others. 
 Consequently, \toolx that gets trained on the aggregate data of all eight domains demonstrates the best results. 

\begin{table}[t]
\centering
\resizebox{\columnwidth}{!}{
\begin{tabular}{p{0.33\columnwidth}>{\centering}p{0.15\columnwidth}>{\centering}p{0.12\columnwidth}>{\centering}p{0.22\columnwidth}>{\centering\arraybackslash}p{0.19\columnwidth}}
\toprule
&\textbf{I.I.D.} & \textbf{CG} & \textbf{Zero-shot} & \textbf{Overall} \\
\midrule
\multicolumn{4}{l}{\textbf{Models with entity linking}}\\
\citet{gu2021beyond} & 58.6 & 40.9 & 51.8 & 51.0 \\
\citet{ye2022rng} & \textbf{86.7} & \textbf{61.7} & \textbf{68.8} & \textbf{71.4} \\
\midrule
\multicolumn{4}{l}{\textbf{Models without entity linking}}\\
BART & 81.1 & 31.6 & 3.6 &  28.1\\
CFQ IR & 86.8 & 46.6 & 5.3 & 34.0 \\
\tool & \textbf{87.4} & \textbf{49.5} & \textbf{9.6} & \textbf{36.9} \\
\bottomrule
\end{tabular}
}
\caption{Validation results on \textsc{GrailQA}'s i.i.d, compositional generalization and zero-shot data splits. The results of two groups of methods (\ie with/without entity linking) are not fully comparable.}
\label{tab:cg_grailqa}
\end{table}

\begin{figure*}[ht]
\centering
\subfloat[Embedding visualization on \textsc{Kqa Pro}.]{\includegraphics[width=0.5\textwidth]{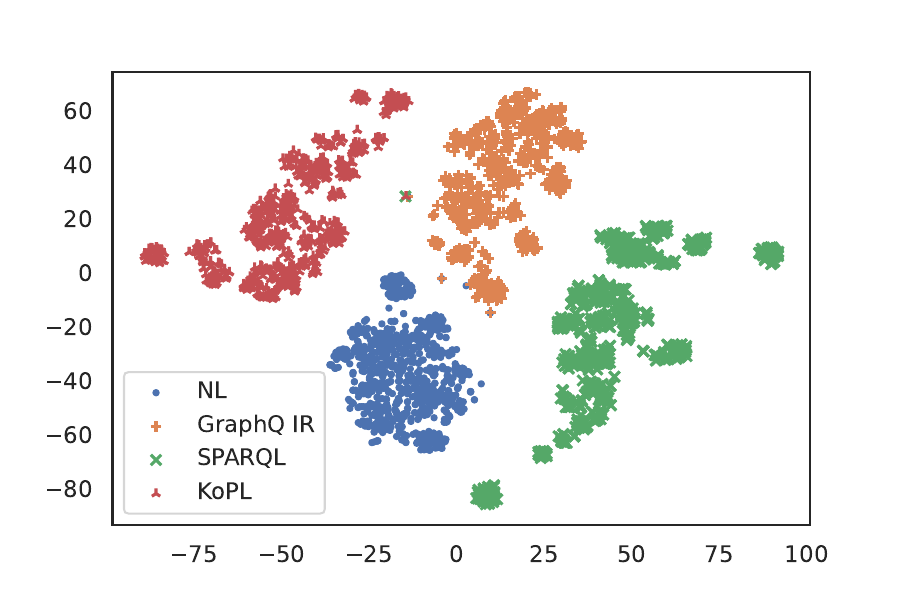}}
\subfloat[Embedding visualization on \textsc{Overnight}.]{\includegraphics[width=0.5\textwidth]{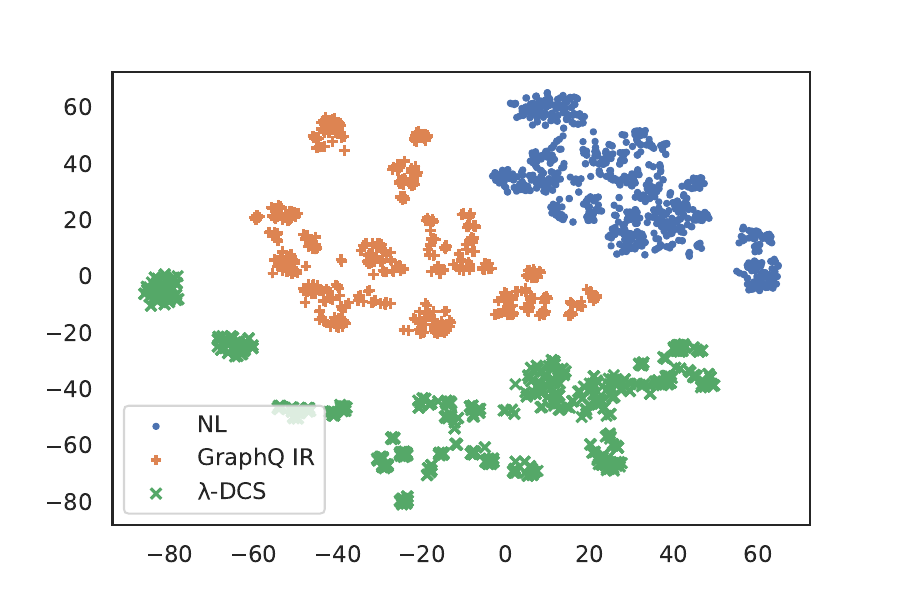}}
\caption{t-SNE visualization of the sequence embeddings of the natural language utterance, \tool and downstream graph query languages that are randomly sampled from the validation set of \textsc{Kqa Pro} and \textsc{Overnight}. }
\label{fig:parameter}
\end{figure*}

\paragraph{OOD Generalization} Current neural semantic parsers often fail in generalizing to out-of-distribution (OOD) data \cite{pasupat2015compositional, keysers2019measuring, furrer2020compositional}. Therefore, we experiment on \textsc{GrailQA}, a dataset that specifically stresses non-i.i.d. generalization. 

We present the results in Table \ref{tab:cg_grailqa}. Among the models without explicit entity linking modules, compared with the BART baseline that directly maps to the logical forms and the CFQ IR \cite{herzig2021unlocking} that particularly aims at SPARQL compositional generalization, \tool achieves the best overall performance and performs remarkably well also in compositional generalization and zero-shot data splits. This can be credited to our IR designs that clarify the redundant semantics and maintain the key hierarchical structure where its components can be flexibly combined or decomposed according to the pre-defined production rules.

\begin{table}[t]
\centering
\resizebox{\columnwidth}{!}{
\begin{tabular}{p{0.35\columnwidth}>{\centering}p{0.25\columnwidth}>{\centering}p{0.25\columnwidth}>{\centering\arraybackslash}p{0.25\columnwidth}}
\toprule
 & \textbf{1-shot} & \textbf{3-shot} & \textbf{5-shot}\\
\midrule
BART & 73.93 & 91.99 & 94.37 \\
\tool & 72.05  & 93.73 & 95.16 \\
\toolx & \textbf{84.91} & \textbf{95.31} & \textbf{96.13} \\
\bottomrule
\end{tabular}
}
\caption{Few-shot learning results on \textsc{MetaQA}-Cypher dataset. \toolx model has formerly trained on \textsc{Kqa Pro} dataset prior to the few-shot fine-tuning.}
\label{tab:metaqa}
\end{table}

\paragraph{Low-resource Generalization} 
To verify whether \tool can aid the semantic parsing of low-resource languages, we reconstruct the \textsc{MetaQA} dataset into Cypher, a graph query language commonly used in the industry but rarely studied in previous semantic parsing works \cite{seifer2019empirical}. To simulate the low-resource scenario, we adjust the data split to ensure that only 1, 3, and 5 samples of each question type appear in the training set under the 1-, 3-, and 5-shot settings. 

The results in Table \ref{tab:metaqa} indicate that our methods can remain robust under a low-resource setting with strong few-shot generalization. Specifically, the \toolx model that has in advance trained on \textsc{Kqa Pro} (a dataset annotated in SPARQL and KoPL) demonstrates the most outstanding performance on \textsc{MetaQA}-Cypher, especially under the most challenging 1-shot setting. Previous works in semantic parsing usually target a specified type of logical form and neglect the data interoperability across languages. With the \tool as a bridge, low-resource query languages can now leverage data from other languages. A universal semantic parser that can end-to-end support different languages also becomes possible.
\begin{table}[t]
\centering
\resizebox{\columnwidth}{!}{
\begin{tabular}{p{0.3\columnwidth}>{}p{0.45\columnwidth}>{\centering\arraybackslash}p{0.35\columnwidth}}
\toprule
& & NL $\Leftrightarrow$ IR \\ 
\midrule
\multirow{2}{*}{\ \ \textsc{Kqa Pro}} & NL $\Leftrightarrow$ SPARQL & -25.28\%  \\
\multicolumn{1}{c}{} & NL $\Leftrightarrow$ KoPL & -16.57\% \\ 
\midrule
\ \ \textsc{Overnight} & NL $\Leftrightarrow$ Lambda-DCS & -15.80\%  \\ 
\bottomrule
\end{tabular}
}
\caption{Semantic distance between natural language utterances and \tool (\ie NL $\Leftrightarrow$ IR) relatively compared to the distance between natural language utterances and specified logical forms.}
\label{tab:distance}
\end{table}

\begin{table*}[ht!]
\centering
\resizebox{\textwidth}{!}{
\begin{tabular}{p{0.15\textwidth}>{\centering}p{0.05\textwidth}>{\centering}p{0.075\textwidth}>{\raggedright\arraybackslash}p{1.35\textwidth}}
\toprule
\textbf{Error Type} & \textbf{\#} & \textbf{\# OSC} & \textbf{Example} \\
\midrule
\multirow{4}[25]{0.15\textwidth}{Inaccurate data annotation} & \multirow{4}[25]{*}{28} & \multirow{4}[25]{*}{-} & \textbf{User utterance:} Out of newscasts that last 110 minutes, which is the shortest? \\
\cline{4-4}
&&& \textbf{Gold SPARQL:} \texttt{SELECT ?e WHERE \{ ?e instance\_of ?c . ?c name "newscast" . ?e duration ?pv\_1 . ?pv\_1 unit "minute" . \red{?pv\_1 value ?v\_1 . FILTER ( ?v\_1 != "110"\string^\string^xsd:double )} . ?e duration ?pv . ?pv value ?v \} ORDER BY ?v LIMIT 1} \\
\cline{4-4}
&&& \textbf{Generated IR:} \texttt{which one has the smallest <A> duration </A> among <ES> <C> newscast </C> whose <A> duration </A> \red{is number <V> 110 minute </V>} </ES>} \\
\cline{4-4}
&&& \textbf{Compiled SPARQL:} \texttt{SELECT ?e WHERE \{ ?e instance\_of ?c . ?c name "newscast" . ?e duration ?pv\_1 . ?pv\_1 unit "minute" . \red{?pv\_1 value "110"\string^\string^xsd:double} . ?e duration ?pv . ?pv value ?v \} ORDER BY ?v LIMIT 1} \\

\hline
\multirow{4}[15]{0.15\textwidth}{Ambiguous query expression} & \multirow{4}[15]{*}{27} & \multirow{4}[15]{*}{27} & \textbf{User utterance:} How is the kid's movie The Spiderwick Chronicles related to John Sayles? \\
\cline{4-4}
&&& \textbf{Gold SPARQL:} \texttt{SELECT DISTINCT ?p WHERE \{ ?e\_1 name "The Spiderwick Chronicles" . ?e\_1 genre ?e\_3 . \red{?e\_3 name "children's film"} . ?e\_2 name "John Sayles" . ?e\_1 ?p ?e\_2 \}} \\
\cline{4-4}
&&& \textbf{Generated IR:} \texttt{what is the relation from <ES> <E> The Spiderwick Chronicles </E> (<ES> ones that <R> genre </R> backward to \red{<E> kid film </E>} </ES>) </ES> to <E> John Sayles </E>} \\
\cline{4-4}
&&& \textbf{Compiled SPARQL:} \texttt{SELECT DISTINCT ?p WHERE \{ ?e\_1 name "The Spiderwick Chronicles" . ?e\_1 genre ?e\_3 . \red{?e\_3 name "kid film"} . ?e\_2 name "John Sayles" . ?e\_1 ?p ?e\_2 \}} \\

\hline
\multirow{4}[15]{0.15\textwidth}{Unspecified graph structure} & \multirow{4}[15]{*}{13} & \multirow{4}[15]{*}{9} & \textbf{User utterance:} When did Tashkent become the capital of Uzbekistan? \\
\cline{4-4}
&&& \textbf{Gold SPARQL:} \texttt{SELECT DISTINCT ?qpv WHERE \{ \red{?e\_1 name "Tashkent" . ?e\_2 name "Uzbekistan" . ?e\_1 capital\_of ?e\_2} . [ fact\_h ?e\_1 ; fact\_r capital\_of ; fact\_t> ?e\_2 ] start\_time ?qpv \}} \\
\cline{4-4}
&&& \textbf{Generated IR:} \texttt{what is the qualifier <Q> start time </Q> of \red{<E> Uzbekistan </E> that <R> capital </R> to <E> Tashkent </E>}} \\
\cline{4-4}
&&& \textbf{Compiled SPARQL:} \texttt{SELECT DISTINCT ?qpv WHERE \{ \red{?e\_1 name "Uzbekistan" . ?e\_2 name "Tashkent" . ?e\_1 capital ?e\_2} . [ fact\_h ?e\_1 ; fact\_r capital ; fact\_t ?e\_2 ] start\_time ?qpv \}} \\

\hline
\multirow{4}[25]{0.15\textwidth}{Nonequivalent semantics} & \multirow{4}[25]{*}{32} & \multirow{4}[25]{*}{25} & \textbf{User utterance:} When did Joseph L. Mankiewicz graduate from Columbia University? \\
\cline{4-4}
&&& \textbf{Gold SPARQL:} \texttt{SELECT DISTINCT ?qpv WHERE \{ ?e\_1 name "Joseph L. Mankiewicz" . ?e\_2 name "Columbia University" . ?e\_1 educated\_at ?e\_2 . \red{[ fact\_h ?e\_1 ; fact\_r educated\_at ; fact\_t ?e\_2 ] end\_time ?qpv} \}} \\
\cline{4-4}
&&& \textbf{Generated IR:} \texttt{what is the qualifier \red{<Q> start time </Q>} of <E> Joseph L. Mankiewicz </E> that <R> educated at </R> to <E> Columbia University </E>} \\
\cline{4-4}
&&& \textbf{Compiled SPARQL:} \texttt{SELECT DISTINCT ?qpv WHERE \{ ?e\_1 name "Joseph L. Mankiewicz" . ?e\_2 name "Columbia University" . ?e\_1 educated\_at ?e\_2 . \red{[ fact\_h ?e\_1 ; fact\_r educated\_at ; fact\_t ?e\_2 ] start\_time ?qpv} \}} \\
\bottomrule
\end{tabular}
} 
\caption{The analysis of 4 error types based on the failure cases as occurred in benchmark \textsc{Kqa Pro}'s test data. ``\# OSC'' refers to the number of errors that can be fixed with one step correction on the IR's structure. }
\label{tab:error}
\end{table*}

\section{Discussion}
To further explore the reasons behind the superior performance of our methods, we compute and visualize the semantic distance between the natural language utterances and their corresponding logical forms or \tool. 

Specifically, to simulate how a neural semantic parser processes the sequences in the above experiments, we use a pretrained BART-base model without fine-tuning to obtain the contextualized embeddings \cite{DBLP:conf/emnlp/LiZHWYL20}. For each sequence, we take the average of the encoder outputs across all word tokens to obtain a 768-dimensional vector as its sentence embedding \cite{ni2022sentence}. Thereafter, we measure the semantic distance between two sequences by computing the Euclidean distance (L2-norm) of their embeddings \cite{chandrasekaran2021evolution}.

% \begin{equation}
%     d(a,b) = \sqrt{\sum^{N}_{i=1}(a_i-b_i)^2},
% \end{equation}
% where $a,b$ are two $N$-dimensional vectors with $a_i$ and $b_i$ denotes the $i$-th value of the vector.

We randomly sampled 1000 queries respectively from \textsc{Kqa Pro} and \textsc{Overnight}'s validation set. We compare the semantic distance between natural language utterances and the GraphQ IRs (\ie NL $\Leftrightarrow$ IR), as well as the distance between natural language utterances and their corresponding logical forms (\eg NL $\Leftrightarrow$ SPARQL). 

The results are listed in Table \ref{tab:distance}. The semantic distance from natural language utterances to \tool is significantly closer than that to different logical forms by at most 25.28\%. We also use t-SNE \cite{van2008visualizing} to reduce the dimension and visualize the embeddings. Figure \ref{fig:parameter} (a) and (b) respectively shows the visualized feature space on \textsc{Kqa Pro} and \textsc{Overnight} datasets. The computation and visualization results affirm our hypothesis that \tool can effectively close the semantic gap and ease the learning of neural semantic parser.

\subsection{Error Analysis}
\label{sec:error}
To investigate \tools potentials and bottlenecks, we look into the failures of our approach when incorrect logical forms are generated. 
Out of the total 979 errors in \textsc{Kqa Pro}'s test set, we randomly sampled 100 cases and categorized them into 4 types as shown in Table \ref{tab:error}.

\paragraph{Inaccurate data annotation} (28\%). The reference logical form (\eg \texttt{v\_1 != "110"}) may contain inconsistent or misinterpreted information that contradicts to the corresponding natural language utterance (\eg \textit{ last 110 minutes}). We attribute this type of error to the dataset rather than the failure of our approach. 

\paragraph{Ambiguous query expression} (27\%). The semantics of the user utterance may be present in more than one way (\eg \textit{kid film} or \textit{children's film}) due to the ambiguity in natural language, whereas the schema of the knowledge base is pre-defined (\eg only \texttt{children's film} is considered a valid entity). This category of error can be fixed by incorporating explicit schema linking modules, which are orthogonal to the implementation of our \tool and semantic parser.

\paragraph{Unspecified graph structure} (13\%). Logical forms of different structures (\eg \texttt{(Uzbekistan capital Tashkent)} and \texttt{(Tashkent capital\_of Uzbekistan)}) can convey the same semantics in a directed cycle graph, but some of them contain structures that are absent in a knowledge base. This type of error is due to the incompleteness of the knowledge base.

\paragraph{Nonequivalent semantics} (32\%). The output includes incorrect query element (\eg string and numerical values) or structure (\eg edges and properties) that conveys nonequivalent semantics, such as misinterpreting \textit{graduate} to \texttt{start\_time}. 

Overall, 89\% of the sampled errors can be simply fixed by the revision of annotation or one-step correction on the IR element, demonstrating that our proposed method with \tool can generate high-quality logical forms that are easy to debug.
\section{Related Work}
\subsection{Semantic Parsing}
Semantic parsing aims to translate natural language utterances into executable logical forms, such as  CCG~\cite{zettlemoyer2005learning}, Lambda-DCS~\cite{liang2013lambda,pasupat2015compositional}, SQL~\cite{zhong2017seq2sql, yu2020grappa}, AMR~\cite{banarescu2013abstract}, SPARQL~\cite{sun2020sparqa} and KoPL~\cite{shi2020kqa,Cao2021ProgramTF}.

Most recent works take semantic parsing as a Seq2Seq translation task via an encoder-decoder framework, which is challenging due to the semantic and structural gaps between natural utterances and logical forms. To overcome such issues, current semantic parsers usually (1) rely on a large amount of labeled data~\cite{shi2020kqa}; or (2) leverage external resources for mini the structural mismatch, \eg injecting grammar rules during decoding~\cite{Wu2021FromPT,shin2021constrained}; or (3) employ synthetic data to diminish the semantic mismatch~\cite{xu2020autoqa, Wu2021FromPT}.

Compared with previous works, our proposed \tool allows the semantic parser to adapt to different downstream formal query languages without extra efforts and demonstrates promising performance under the compositional generalization and few-shot settings.

\subsection{Intermediate Representation}
Intermediate representations (IR) are usually generated for the internal use of compilers and represent the code structure of input programs \cite{aho2007compilers}. Good IR designs with informative and distinctive mid-level features can provide huge benefits for optimization, translation, and downstream code generation \cite{lattner2004llvm}, especially in areas like deep learning \cite{chen2018tvm, cyphers2018intel} and heterogeneous computing \cite{lattner2020mlir}. 

Recently, IR has also become common in many semantic parsing works that include an auxiliary representation between natural language and logical form. Most of them take a top-down approach and adopt IR similar to natural language \cite{su2017cross, herzig2019don, shin2021constrained}. In contrast, another category of works constructs IR based on the key structure of target logical forms in a bottom-up manner \cite{wolfson2020break, marion2021structured}. For example, \citeauthor{herzig2021unlocking} designed CFQ IR that rewrites SPARQL by grouping the triples of identical elements \citeyearpar{herzig2021unlocking}. 

Although these works partially mitigate the mismatch between natural and formal language, they either failed in removing the formal representations that are unnatural to the language models or neglected the structural information requisite for downstream compilation. In this work, we omit those IRs that cannot be losslessly converted into downstream logical forms.

% In contrast, \tool can benefit from both approaches with its natural-language-like semantic representation and formal-language-like syntactic definition. The unification also saves the huge costs incurred when preparing separate IR and compiler for each graph query language in previous works.

% \subsection{Knowledge Base Question Answering} 
% For natural language question-answering over knowledge bases, current methods can be mainly grouped into two categories: 1) semantic parsing methods, as described above; and 2) information retrieval methods~\cite{miller2016key,saxena2020improving,schlichtkrull2018rgcn,zhou2018interpretable,qiu2020stepwise} that construct a question-specific graph extracted from the knowledge base and rank all the entities in the extracted graph based on their relevance to the question. Compared with information retrieval methods, semantic-parsing-based methods can provide better interpretability by generating expressive logic forms.
\section{Conclusion and Future Work}
This paper proposes a novel intermediate representation, namely \tool, for bridging the semantic gap between natural language and graph query languages. Evaluation results show that our approach with \tool consistently surpasses the baselines on several benchmarks covering multiple formal languages, \ie SPARQL, KoPL, Lambda-DCS, and Cypher. Moreover, \tool also demonstrates superior generalization ability and robustness under the out-of-distribution and low-resource settings. 

As an early step towards the unification of semantic parsing, our work opens up several future directions. For example, many code optimization techniques (\eg common subexpression elimination) can be incorporated into IR to improve performance further. By bringing in multiple levels of IR, our framework may also be extended to support relational database query languages like SQL. Moreover, since the current designs of \tool still require non-trivial manual efforts, the automation of such procedure, \eg in prompt-like manners, is worth future exploration.

\section*{Limitations}
The major limitations of this work include: (a) the composition rules of \tool are closely aligned with interrogative sentences. Therefore, our current formalism may not be applicable to general-domain semantic parsing; (b) for the semantic parsing of an input language whose syntax significantly differs from English (\eg Arabic, Chinese, Hindi, etc.), the benefits of \tool may be limited; (c) our experiments fine-tuned a neural semantic parser on top of a pretrained model with $\sim$139 million parameters, thus cannot be easily reproduced without adequate GPU resources. 

\section*{Acknowledgements}
We would like to thank the anonymous reviewers for their valuable comments. This work is partially supported by the National Key R\&D Program of China (2021ZD0110104), the National Natural Science Foundation of China (U20A20226), the NSFC Youth Project (62006136), and a grant from the Institute for Guo Qiang, Tsinghua University (2019GQB0003). Jidong Zhai is the corresponding author of this paper.

\bibliography{anthology,ref}
\bibliographystyle{acl_natbib}

\newpage
\appendix
\begin{table*}[!ht]
\small
\centering
\resizebox{\textwidth}{!}{
\begin{tabular}{p{0.25\textwidth}>{\raggedright\arraybackslash}p{0.9\textwidth}}
\toprule
\textbf{Non-terminal} & \textbf{Productions}  \\
\midrule
$S$ &$\rightarrow$ \texttt{EntityQuery} | \texttt{AttributeQuery} | \texttt{RelationQuery} |  \texttt{QualifierQuery} |  \texttt{CountQuery} | \texttt{VerifyQuery} | \texttt{ValueQuery}  \\
\hline
\texttt{EntityQuery} &$\rightarrow$  \textit{what is} \texttt{EntitySet}  \\
\hline
\texttt{AttributeQuery} &$\rightarrow$  \textit{what is the attribute} \texttt{Attribute} \textit{of} \texttt{EntitySet}  \\
\hline
\texttt{RelationQuery} &$\rightarrow$ \textit{what is the relation from} \texttt{EntitySet} \textit{to} \texttt{EntitySet} \\
\hline
\texttt{QualifierQuery} &$\rightarrow$  \textit{what is the qualifier} \texttt{Qualifier} \textit{of} \texttt{EntitySet Constraint}  \\
\hline
\texttt{CountQuery} &$\rightarrow$  \textit{how many} \texttt{EntitySet}  \\
\hline
\texttt{VerifyQuery} &$\rightarrow$  \textit{whether} \texttt{EntitySet Constraint}  \\
\hline
\texttt{ValueQuery} &$\rightarrow$  \textit{what is} \texttt{Value}  \\
\hline
\texttt{EntitySet} &$\rightarrow$ \texttt{<ES> EntitySet LOP EntitySet </ES>} | \texttt{<ES>} \texttt{EntitySet Constraint} \texttt{</ES>} | \texttt{<ES> Concept EntitySet </ES>} | \texttt{Concept} | \texttt{Entity} | \textit{ones} \\
\hline
\texttt{Constraint} &$\rightarrow$ \texttt{AttributeConstraint} \texttt{QualifierConstraint?} | \texttt{RelationConstraint} \texttt{QualifierConstraint?} \\
\hline
\texttt{AttributeConstraint} &$\rightarrow$ \textit{whose} \texttt{Attribute COP Value} | \textit{that have} \texttt{SOP} \texttt{Attribute} \\
\hline
\texttt{RelationConstraint} &$\rightarrow$ \textit{that} \texttt{Relation DIR} \textit{to} (\texttt{COP Value?}) \texttt{EntitySet} | \textit{that} \texttt{Relation DIR} \textit{to} \texttt{SOP} \texttt{EntitySet}  \\
\hline
\texttt{QualifierConstraint} &$\rightarrow$ \texttt{Qualifier} \texttt{COP Value}\\
\hline
\texttt{Entity} &$\rightarrow$ \texttt{<E>} \textit{entity} \texttt{</E>}  \\
\hline
\texttt{Concept} &$\rightarrow$ \texttt{<C>} \textit{concept} \texttt{</C>} \\
\hline
\texttt{Attribute} &$\rightarrow$ \texttt{<A>} \textit{attribute} \texttt{</A>}\\
\hline
\texttt{Relation} &$\rightarrow$ \texttt{<R>} \textit{relation} \texttt{</R>} \\
\hline
\texttt{Qualifier} &$\rightarrow$ \texttt{<Q>} \textit{qualifier} \texttt{</Q>} \\
\hline
\texttt{Value} &$\rightarrow$  \texttt{VTYPE <V>} \texttt{Value LOP Value} | \texttt{VOP} \textit{of} \texttt{Value} | \texttt{Attribute} \textit{of} \texttt{Entity} | \texttt{VTYPE <V>} \textit{value} \texttt{</V>} \\
\hline
\texttt{LOP} &$\rightarrow$ \textit{and} | \textit{or} | \textit{not}  \\
\hline
\texttt{VOP} &$\rightarrow$ \textit{sum} | \textit{average} | \textit{maximum} | \textit{minimum} \\
\hline
\texttt{COP} &$\rightarrow$ \textit{is} | \textit{is not} | \textit{larger than} | \textit{smaller than} | \textit{at least} | \textit{at most} \\
\hline
\texttt{SOP} &$\rightarrow$ \textit{largest} | \textit{smallest} \\
\hline
\texttt{DIR} &$\rightarrow$ \textit{forward} | \textit{backward}  \\
\hline
\texttt{VTYPE} &$\rightarrow$ \textit{string} | \textit{numeric} | \textit{year} | \textit{month} |  \textit{date} | \textit{time}  \\
\bottomrule
\end{tabular}
} 
\caption{\tool grammar rules that cover the common graph query patterns. ``\texttt{|}'' separates multiple productions at the same level, and ``\texttt{?}'' denotes that the preceding expression is optional. Italic words refer to the terminal symbols. Here we omit the corner case production rules for simplicity.}
\label{tab:full_grammar}
\end{table*}
\section{\tool Grammar}
We present \tools non-terminals and production rules in Table \ref{tab:full_grammar}. 

\section{Implementation Details}
\subsection{Model Hyperparameters}

For the neural semantic parser, we used the BART-base model \cite{lewis2020bart} released by Facebook on HuggingFace\footnote{https://huggingface.co/facebook/bart-base}. 12 special tokens (\eg \texttt{<E>}) were added to the tokenizer vocabulary as the structure indicators for \tool. We used the AdamW optimizer \cite{loshchilov2018fixing} with the learning rate set to $3e^{-5}$ and weight decay set to $1e^{-5}$ following the default settings. 

\subsection{Environmental Configurations}

In our implementation of the compiler, we used ANTLR \cite{parr2013definitive} version 4.9.2 for analyzing our specified grammar rules and building up the corresponding lexer and parser toolkit. For evaluation, we used Virtuoso 7.20\footnote{https://github.com/openlink/virtuoso-opensource}, SEMPRE 2.4\footnote{https://github.com/percyliang/sempre}, Neo4j 4.4\footnote{https://github.com/neo4j/neo4j} and KoPL 0.3\footnote{https://pypi.org/project/KoPL/} as the back-ends respectively for executing the SPARQL, Lambda-DCS, Cypher, and KoPL queries. 

Our whole experiments were performed on a single machine with 8 NVIDIA Tesla V100 (32GB memory) GPUs on CUDA 11. 

\section{Supplementary Study}
\subsection{\textsc{Kqa Pro} Compositional Generalization}

\begin{table}[t]
\centering
\small
\resizebox{\columnwidth}{!}{
\begin{tabular}{lccccc}
\toprule
& \textbf{Overall} &\textbf{Qualifier} & \textbf{Comparison} &\textbf{Logical} \\
\midrule
BART & 50.58 & 21.55 & 87.66 & 50.60 \\
CFQ IR & 50.70 & 25.33 & 93.77 & 50.73  \\
\tool & \textbf{54.91} & \textbf{40.46} & \textbf{95.19} & \textbf{54.90} \\
\bottomrule
\end{tabular}
}
\caption{Experimental results on \textsc{Kqa Pro} compositional generalization data split.}
\label{tab:cg_kqapro}
\end{table}

Compositional generalization refers to a model's capability of generalizing from the known components to produce novel combinations \cite{pasupat2015compositional, keysers2019measuring, furrer2020compositional}. To measure our IR's compositional generalization ability, we also create a new \textsc{Kqa Pro} data split based on the logical form length and test the parsers to generate long queries (KoPL queries with $>$ 7 functions) based on the short query components seen in the training data  (KoPL queries with $\leq$ 7 functions).  

The results are listed in Table \ref{tab:cg_kqapro}. Compared with the plain-BART baseline and the CFQ IR \cite{herzig2021unlocking} that is specially designed for improving the compositional generalization on SPARQL, \tool achieves the best performance in overall data as well as in complex task settings, which can be again credited to our IR designs that simplify the redundant semantics and preserve the key structural features. 

\begin{figure*}[t]
    \centering
    \resizebox{.93\linewidth}{!}{
    \includegraphics{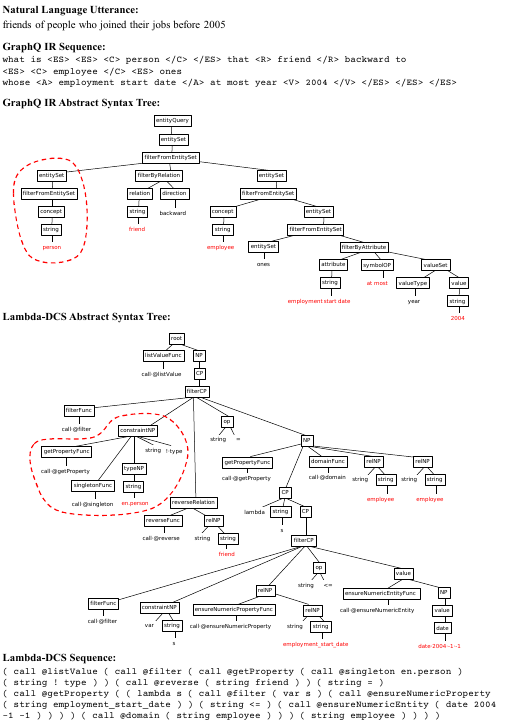}
    }
    \caption{A user query in \textsc{Overnight}. The neural semantic parser first converts the input utterance into \tool. The compiler then parses the \tool sequence into an abstract syntax tree, which is subsequently transformed into the corresponding Lambda-DCS sequence along with a tree mapping process. To exemplify, the subtrees circled by red dash lines are carrying equivalent information that can be transformed with pre-defined rules. The red words are terminal nodes that correspond to the graph structure. }
    \label{fig:ast_1}
\end{figure*}

\begin{figure*}[t]
    \centering
    \resizebox{.93\linewidth}{!}{
    \includegraphics{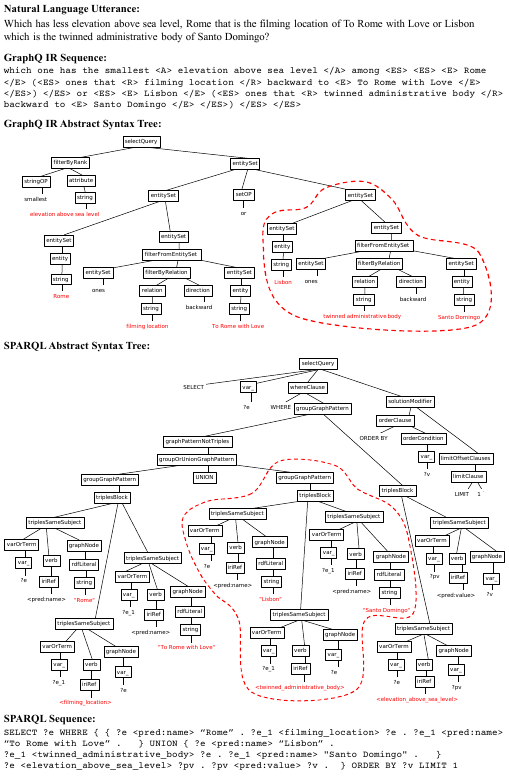}
    }
    \caption{A user query in \textsc{KQA Pro}. Similarly, the compiler parses the generated \tool sequence into an abstract syntax tree, then transform its tree structure into the corresponding SPARQL sequence. }
    \label{fig:ast_2}
\end{figure*}

\end{document}